\documentclass[11pt,a4paper]{article}
\usepackage[hyperref]{acl2020}
\usepackage{times}
\usepackage{latexsym}
\usepackage{xcolor,colortbl}
\usepackage{times}
\usepackage{graphicx}
\usepackage{color}
\usepackage{multicol}
\usepackage{url}
\usepackage{multirow}
\usepackage{amssymb}
\usepackage{tabularx}
\usepackage{booktabs}
\usepackage{comment}

\usepackage{microtype}

\aclfinalcopy

\title{Implicit Knowledge in Argumentative Texts: An Annotated Corpus}

\author{Maria Becker, Katharina Korfhage, Anette Frank \\ Institute of Computational Linguistics Heidelberg University \\
         (mbecker|korfhage|frank)@cl.uni-heidelberg.de\\}

\begin{document}
\maketitle
\begin{abstract}
When speaking or writing, people omit information that seems clear and evident, such that only part of the message is expressed in words. Especially in argumentative texts it is very common that (important) parts of the argument are implied and omitted. We hypothesize that 
% Starting from the observation that/entymeme recosntruction, but: no existing corpora
for argument analysis it will be beneficial to reconstruct this implied information. As a starting point for filling such knowledge gaps, we build a corpus consisting of high-quality human annotations of missing and implied information in argumentative texts. To learn more about the characteristics of both the argumentative texts and the added information, 
%more concrete
we further annotate the data with 
%semantic information
semantic clause types and commonsense knowledge relations. The outcome of our work is a carefully designed and richly annotated dataset, for which we then provide an in-depth analysis by investigating characteristic distributions and correlations of the assigned labels. We reveal interesting patterns and intersections between the annotation categories and properties of our dataset, which enable insights into the characteristics of both argumentative texts and implicit knowledge in terms of structural features and semantic information. The results of our analysis can help to assist automated argument analysis and can guide the process of revealing implicit information in argumentative texts automatically. %We then finally suggest use cases and downstream applications for which our dataset can be leveraged.
\end{abstract}
%\\ \newline \Keywords{implicit knowledge, argumentation, annotation, semantic clause types, commonsense knowledge relations, ConceptNet} }

%\maketitleabstract

\section{Introduction}

In everyday communication as well as in written texts people omit information that seems clear and evident, such that only part of the message needs to be expressed in words \cite{grice1975logic}. While this information can easily be filled in by the hearer,  a computational system typically does not possess commonsense or domain-specific knowledge that is needed to reconstruct the implied information. Especially in argumentative texts it is very common that (important) parts of the argument such as warrants are implied and omitted \cite{Rajendranetal:16, beckeretal:2017a, hulpus19}. This leads us to the assumption that the logic of an argument is in general not fully recoverable from what is explicitly said, and that for argument analysis it will be beneficial to reconstruct such implied information.\\
We aim to fill such gaps by identifying and inserting knowledge that connects given statements. To perform this, we want to learn from human-generated data of missing and implied information. This motivates the current work, in which we gather high-quality annotations of implied knowledge in the form of simple natural language sentences. The annotations are performed on pairs of argumentative units from the Microtexts Corpus \cite{peld:stede:2015}, a very concise and focused argumentation dataset which is already annotated with argumentative components and relations such as \textit{support, rebuttal} or \textit{undercut}. For all unit pairs they are presented with, annotators are asked to add the information that makes the connection between the units explicit, using short and simple sentences. To learn more about the nature and %(linguistic) 
characteristics of both the argumentative texts and the added information, we further annotate the data with two specific semantic information types: semantic clause types \cite{friedrich2014automatic} and ConceptNet knowledge relations \cite{Speeretal:2012,Havasietal:2009}, which were both found to be characteristic for argumentative texts \cite{beckeretal:2016a,beckeretal:2017a}.
The outcome of our work is a carefully designed and richly annotated dataset, for which we provide an in-depth analysis by investigating characteristic distributions and correlations between the assigned labels. 

The contributions of this work are: (i) high-quality annotations of implicit knowledge on the argumentative Microtext corpus, (ii) characterization of the argumentative units from the Microtext corpus and the inserted sentences in terms of semantic clause types and commonsense knowledge relations; and (iii) an in-depth study of properties and correlations of the assigned labels. The dataset will be made public as an extension to the Microtext corpus \cite{peld:stede:2015} to support further research in argument analysis.
\section{Related Work}

\textbf{Finding and Adding Implicit Knowledge in Arguments.} Relatively little attention has been devoted so far to the task of finding and adding implicit knowledge in arguments, which is closely related to the task of \textbf{enthymeme} reconstruction. Enthymemes -- arguments with missing propositions -- are common in natural language and particularly in argumentative texts \cite{Rajendranetal:16}. \newcite{Razuvayevskayaetal:16} present a feasibility study on the automatic detection of enthymemes in real-world texts and find that specific discourse markers (e.g. {\em let alone, %therefore, 
because}) can signal enthymemes. Using these as trigger words, they reconstruct enthymemes from the local context, while \newcite{Rajendranetal:16} retrieve and fill missing propositions in arguments from similar or related arguments. %Another used method is the utilization of shared knowledge \cite{BlackHunter:2012}, which is related to our approach. 
\\
\newcite{beckeretal:2016a} and \newcite{beckeretal:2016b} show that argumentative texts are rich in \textbf{generic and generalizing sentences}, which are semantic clause types \cite{friedrich2014automatic} that often express commonsense knowledge. We will show that large portions of implied knowledge in argumentative texts are naturally stated using these clause types. \\
In their attempt to reconstruct implicit knowledge, \newcite{boltuzic2016fill} find that the \textbf{claims} that users make in \textbf{online debate} platforms often build on implicit knowledge. They show that the amount of implicitness is dependent on genre and register and point out that the reconstruction of implicit premises can be helpful for claim detection. \\
Recently, \newcite{hulpus19} point out the relevance of reconstructing implicit knowledge for understanding %and critically assessing 
arguments in a computational setting, by proposing the task of \textbf{argument explicitation}, which they define as a task that makes explicit both (i) the structure of a natural language argument, as well as (ii) the background knowledge the argument is built on, in the form of implicit premises or contextual knowledge. %They discuss specific kinds of knowledge that are needed for argument explicitation such as knowledge about natural language, knowledge about argumentation, and background knowledge, and distinguish between missing premises that are subjective as opposed to those that are facts. 

These studies reinforce the view that a substantial amount of knowledge is needed for the correct interpretation and analysis of argumentative texts, and thus filling knowledge gaps in argumentative texts will be beneficial for argument analysis. %(cf. \cite{Rajendranetal:16}).
%In this work we report on an annotation process through which we acquire high-quality annotations for implicit knowledge in arguments.
\\

\textbf{Related Datasets.} \newcite{boltuzic2016fill} release a small dataset with human-provided \textbf{implicit premises} based on data from \textbf{online debate platforms}, consisting of 125 claim pairs annotated with the premises that connect them. %, yielding a total of 500 gap-filling premises set. 
In contrast to our approach they asked the annotators to provide the premises that bridge the gap between the two claims without giving any further instructions, resulting in a substantial variance in both the wording and the average number of premises.% as well as a low word overlap.% (32\%). 

\newcite{beckeretal:2017a} design a process for obtaining high-quality \textbf{implied knowledge annotations} for \textbf{German argumentative microtexts} \cite{peld:stede:2015}, in the form of simple natural language statements which are then further characterized with semantic clause types and commonsense knowledge relations. % annotations. 
Since the decision of what exactly is missing and how detailed such information should be can be subjective, they propose several steps to promote agreement among the annotators and monitor its evolution using textual similarity computation. The implicit knowledge annotations we present in this paper are also based on argumentative microtexts \cite{peld:stede:2015}, thus our corpus can be seen as an extension of the corpus published by \newcite{beckeretal:2017a}. The main differences are that (i) our data is in English (as opposed to German), (ii) the semantic clause types and commonsense knowledge relations are not only annotated for the inserted sentences, but are also available for the argumentative texts themselves, (iii) our corpus includes more annotated unit pairs, and that (iv) in our corpus all annotations are conducted by expert annotators. 

\newcite{tubiblio104556} present the \textbf{argument reasoning comprehension task}, where given an argument with a claim and a premise, the goal is to choose the correct implicit warrant from two options. Both warrants are plausible and lexically close, but lead to contradicting claims. They provide a dataset %obtained from a crowdsourcing process 
where Amazon Turkers added warrants for 2k arguments from news comments. As opposed to our dataset, the annotators were only supposed to fill in the gap between a pair of claim and premise, while we consider larger arguments consisting of a claim and \textit{several} premises. Furthermore, we annotate implicit information not only between claim and premises, but between all adjacent argument units and all argument units that stand in a direct argumentative relation (cf. Sec.\ 3.2). The second major difference is that in \newcite{tubiblio104556} the annotators were only asked to add \textit{one} warrant (one sentence) per argument, %(plus the alternative, false warrant), 
while we assume that more than one sentence might be needed to fill a knowledge gap in an argument. 

\begin{comment}
Very recently, \newcite{Bhagavatula19} propose abductive \textbf{natural language inference} as a novel reasoning task in narrative texts. They introduce the ART, a large-scale benchmark dataset based on ROCStories. The dataset consists of about 20k commonsense narrative contexts in the form of sentences that express (partial) observations and 200k explanations formulated as multiple choice questions. The task is then to choose which of the explanation are inferable from the given observations. \bluenote{TBD: compare to our dataset}
\end{comment}

\section{Enriching Argumentative Texts with Implicit Knowledge}

\subsection{General Annotation Procedure}
%\bluenote{Short description of annotation procedure, refer to NLDB paper for details, total: 1/2 page}

The main goal of our annotation project is to uncover and characterize implicit knowledge that connects a given pair of argumentative units.
This overall objective is subdivided into two consecutive annotation tasks: 

\begin{description}
\item[i.] First, we ask the annotators to detect missing knowledge that connects a pair of argumentative units, and to express this knowledge in terms of simple natural language statements.
\vspace{-0.3cm}
\item[ii.]
In the next step (cf. Sec.\ 4), the annotators are tasked with labeling both the inserted sentences and the given argumentative text units with characterizing semantic information. The annotation types that we select are \textit{semantic clause types} \cite{friedrich2014automatic} and \textit{commonsense knowledge relations}, following the ConceptNet relation inventory \cite{Havasietal:2009}.  
\end{description}

\subsection{The Microtext Corpus}

\begin{figure}[t]
\begin{centering}
\includegraphics[width=0.48
\textwidth]{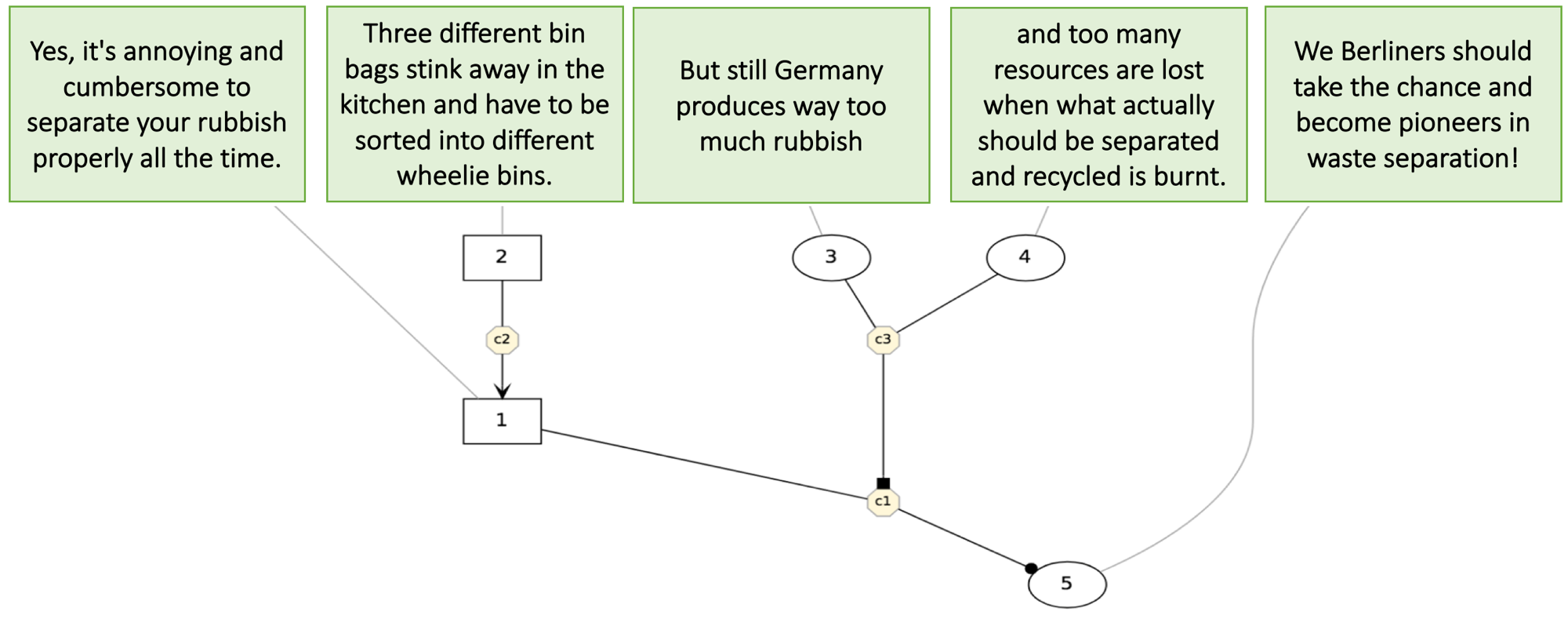}
\vspace{-0.8cm}
\caption{Example of a microtext (argument graph).}
\label{fig:arggraph}
\end{centering}
\end{figure}

As basis for our annotations we use the argumentative Microtext corpus \cite{peld:stede:2015}, which consists of 112 microtexts. The corpus was created in German and has been translated to English. In this work we use only the English version (for annotations on the German version, cf. \newcite{beckeretal:2017a}). Each microtext is a short, dense argument consisting of roughly five elementary units of argumentation, so called argumentative units \cite{peld:stede:2015}.  The texts are written in response to a question on a potentially controversial issue (e.g. \textit{Should there be a cap on rent increases for a change of tenant?}). Writers were asked to include a direct statement of their main claim as well as at least one objection to that claim. The generated arguments were then manually annotated with argumentation graphs (one graph per microtext, cf. Fig.\ \ref{fig:arggraph} for an example) according to a scheme based on Freeman’s theory of the macro-structure of argumentation 
\cite{freeman2011}. % for representing text-level argumentation structure. 
The nodes in the graph are argumentative units and the edges are argumentative relations between them. The most frequent relations are {\em support} (premise which supports a conclusion or another premise), {\em rebuttal} (premise that attacks a conclusion or premise by challenging its acceptability), or {\em undercut} (premise which attacks the acceptability of an argumentative relation between two propositions).\\
For our work we extract pairs of argumentative units that either stand in a direct argumentative relation (that means units that are directly connected in the argument graph), or units that are adjacent to each other, or both. In sum we extract 719 pairs of argumentative units which we then provide to our annotators to perform several different annotation tasks that we describe in the following sections.

%\bluenote{Short description of MC and how we extracted pairs, + one example arg graph, total: 1/2 page}

%\greennote{Hmm, I see that this does not fit your overall section structure; still, I find it important to make this clear at some point. Maybe make this general information section 3.1; put Microtext corpus in 3.2., then have annotation of implicit information 3.3. + statistics 3.4., and then second round of annotations section 4?}

%\greennote{I singled out the no-missing-connection case. You may want to change back to the version where this case is included.}
\subsection{Annotating Implicit Knowledge}
We asked our annotators to detect whether the connection between the pair of units is made fully explicit by the text, and if this is not the case, to explain the missing connection by providing one or more sentences that make this connection explicit. %There may be different situations -- the connection can be obvious or not; it can be a direct link or a more extended connection; the missing information may be expressed in one or more sentences. 
Our annotators were supposed to add as few sentences as possible and to make these sentences very simple (if possible one fact per sentence) in order to retrieve the minimal amount of information that is needed to connect the two units and to avoid too detailed explanations. Since some unit pairs only make sense with a larger context, for every pair we also displayed the full microtext. % to the annotators. 
Figure \ref{fig:annotguidelines} shows two examples of the annotations, where in the first one the main claim \emph{1-a} is attacked by statement \emph{1-b}, while in the second one the premise \emph{2-b} supports the main claim \emph{2-a}. The knowledge underlying the connection between the main claim and the premise in both cases is made explicit in \emph{c} respectively, whereby for the first example one and for the second example two sentences have been inserted. % by our annotators.

\begin{figure}[t]
\noindent (1-a) \noindent\hspace*{1.8mm} \small{\textit{BER should be re-conceptualized from scratch}}\\ 
\noindent (1-b) \noindent\hspace*{3mm} \textit{even if billions of Euros have already been invested}\\ 
\noindent\hspace*{10.5mm}%
\textit{in the existing airport project.}\\
(1-c) \noindent\hspace*{2.5mm} 
\textit{BER is an airport.}\\\\
\noindent (2-a) \noindent\hspace*{3mm} \textit{Capital punishment is not a solution}\\ 
\noindent (2-b) \noindent\hspace*{3mm}  \textit{as it cannot be ruled out that the judicial process}\\ 
\noindent\hspace*{11mm}%
\textit{may make mistakes.}\\ 
(2-c-I) \noindent\hspace*{1mm} 
\textit{In a judicial process it is decided about capital}\\ 
\noindent\hspace*{11mm}%
\textit{punishment.}
\noindent  \ \\
(2-c-II) \noindent\hspace*{0.2mm}
\textit{Mistakes don't lead to solutions.}

\caption{Example annotations for explicating implicit knowledge (c) that connects argumentative units (a\&b).}\label{fig:annotguidelines}
%\vspace{-0.6cm}
\end{figure}

The difficulty of eliciting such implicit knowledge in an annotation task is that intuitions about which knowledge exactly is missing may be different between annotators, and even if their intuitions match, the phrasing may be different, structurally or in terms of lexical choice. In order to enforce agreement and to assess the quality of the annotations, \newcite{beckeretal:2017a} design a multi-step annotation process where annotators are asked to review and revise each other's annotations, whereby the evolution of agreement during this process is monitored using computational measures of semantic textual similarity. \newcite{beckeretal:2017a} use five annotators for each argumentative unit pair, while we train two expert annotators with a linguistic background who produce two versions of the implicit knowledge, which then serve as the basis for the final gold standard produced by another expert annotator (one of the authors). This final adjudicated corpus provides the basis for the second annotation step (Sec.\ 4) and  our analyses in Sec.\ 5.

\subsection{Annotation Statistics}

%\bluenote{TBD: add general statistics (number/length of inserted sentences etc.)}

\textbf{Annotator Agreement.} Building on the insights of \newcite{beckeretal:2017a}, we calculate the semantic similarity of the two initial annotations in order to evaluate the agreement between the annotators and compare it to the similarity scores  reported in \newcite{beckeretal:2017a}. Following \newcite{beckeretal:2017a}, we quantify the distance between the annotations using the Word Mover’s Distance \cite{KusnerSKW15} as implemented in {\it gensim}\footnote{\url{https://radimrehurek.com/gensim}}. The Word Mover’s Distance (WMD) measures the dissimilarity between two documents as the aggregated minimum distance in an embedding space that the (non-stopword) words of one document need to “travel” to reach the (non-stopword) word of another document.
As embeddings, we use 300-dimensional skip-gram word2vec embeddings
trained on part of the Google News dataset (100 billion words, \newcite{Mikolov:2013}).\\
%Our first annotator inserted on average 1.95 sentences and our second annotator 1.77 sentences per argument unit pair. %We add a length difference penalty score to the WMD in order to increase the distance score if there is a large difference in length between the sentences. 
We compare the complete annotation for each argumentative unit pair (as opposed to sentence by sentence) %, for details cf.\ \newcite{beckeretal:2017a} 
and measure a WMD distance score of 1.97.  \newcite{beckeretal:2017a} compare distance scores between implicit knowledge annotations produced in early vs. later stages of their multi-step annotation procedure. In their first two rounds of annotations which include initial annotations and mutual editing and correcting, they compute a WMD of 2.2 and 3.08, and in the third round where the corrected annotations are merged by new annotators, the WMD decreases to 1.89, demonstrating the evolution of annotator agreement. Our score of 1.97 is closest to the score reported for the third round, which we interpret as sufficient agreement between the annotators. %We also compute agreement as measured by word overlap. We obtain an averaged word overlap score of 47.02 (Dice) and 34.03 (Jaccard)...

\section{Annotating Argumentative Texts and Implicit Knowledge with Additional Information}% Semantic Clause Types and Commonsense Knowledge Relations}

\textbf{Learning from Human Annotations.} We hypothesize that the more we know about the knowledge that is needed to establish links between (argumentative) sentences, the easier it will be to reconstruct them automatically. 
All of the following tasks are therefore designed with the ultimate goal of learning more about the properties of the sentences that were stated by our annotators to make the missing information explicit, \textit{within} their surrounding explicit context.\\
We expect semantic clause types to be useful features for characterizing argumentative texts, implicit knowledge \textit{and} their interaction, since clause types have shown to be relevant for interpreting semantics at the clause
level and discourse structure (cf. \newcite{friedrich2014automatic}). Furthermore,  \newcite{beckeretal:2016b} showed that the distribution of these clause types is distinctive for argumentative texts compared to other genres in terms of particularly high ratios of generic and generalizing sentences. \\
We furthermore expect ConceptNet \cite{Speeretal:2012} to be a useful resource for finding and characterizing implicit sentences, since implied information is usually commonsense knowledge that seems clear and evident and is for that reason omitted. ConceptNet provides exactly that kind of information, since it contains commonsense facts about the world and everyday life (cf. Sec. 4.2). Also, the relation inventory of ConceptNet is targeted for capturing commonsense knowledge, and we therefore expect it to be appropriate for labeling and characterizing implicit knowledge.

What additionally makes clause types and commonsense relations attractive features for analyzing and characterizing argumentative texts and implicit knowledge is that recently for both -- semantic clause types (\newcite{beckeretal:2017c}) and commonsense relations (\newcite{beckeretal:2019}) -- automated classification models have been published, which can be used for pre-labeling the given texts and therefore facilitate the automatic analysis of arguments and implicit knowledge. % and their semantic properties. 

\subsection{Annotating Semantic Clause Types}

\textbf {Inventory and Annotation Process.} 
We asked the annotators to characterize both the argumentative units from the microtexts and the gold standard of the inserted sentences by labeling them with {\em Semantic Clause Types}. For the inventory %of semantic clause types 
we adopt the most frequent types in \newcite{friedrich2014automatic} and give examples from our dataset:

\begin{description}
\itemsep-4.5pt
 \item[States] (STA) describe specific properties of individuals: \\
\textit{The Mayor of Berlin has an interest in Berlin's coffers.}
 \item[Events] (EVT) are things that happen or have happened: \\
\textit{Edward Snowden revealed information.} 
 \item[Generic Sentences] (GEN) are predicates over classes or kinds:
\textit{Supermarkets should open on Sundays.}
 \item[Generalizing Sentences] (GNZ) describe regularly occurring events/habits:
\textit{Germany produces much rubbish.}
\end{description}

The annotations are performed independently by two trained annotators who assign labels at the clause level, whereby one sentence may contain more than one clause. %The gold standard is produced by an expert annotator (one of the authors) and provides the basis of our analysis. 
\\

\textbf{Statistics.} We measure a fair annotator agreement of 34.02\% (Cohen's Kappa) and produce a gold standard done by an expert annotator (one of the authors) that provides the basis of our final analysis. Table \ref{tab:SE_dis} displays the distribution of semantic clause types within the implicit information  annotations and the argumentative units from the microtexts, which we then compare to the numbers reported for other genres \cite{beckeretal:2016b}. We find a high proportion of \textsc{Generics} within the Microtexts (64\%) and an even higher amount within he implicit information annotations (84\%), while the other genres (reports, speeches, fiction) rather contain mostly \textsc{States} and \textsc{Events}. This indicates the relevance of knowledge captured by \textsc{Generic Sentences} within the added implicit information, and we can use this finding for acquiring such missing information automatically.

%\greennote{correlation between +/-implicitness for specific argumentative relations and/or adjacency}

{\small 
\begin{table}[t]\begin{tabular}{ p{2.9cm}p{0.9cm}p{0.9cm}p{0.9cm}p{0.9cm} }
%	 \hline
	 \toprule
	 \textbf{Genre} & {\bf GEN}&{\bf GNZ}& {\bf STA}& {\bf EVT}\\ \midrule
%	 \hline
	Impl. Information
 &0.84&0.02&0.13&0.01\\
Microtexts &0.64&0.05&0.24&0.02\\ 
\hline
Report &0.03&0.04&0.54&0.39\\ 
TED Talk &0.12&0.03&0.49&0.36\\ 
Fiction &0.02&0.05&0.39&0.54\\ 
%	 \hline
\bottomrule
    \end{tabular}
    \caption{Distribution of the most frequent Semantic Clause Types among different genres (expressed as percentages)}
\label{tab:SE_dis}
\vspace{0.3cm}
    \end{table}
    }

\subsection{Annotating Commonsense Knowledge Relations}

\textbf {Inventory and Annotation Process.} In addition to clause types we annotate the argumentative units and the %gold standard of the 
inserted sentences with ConceptNet relation types. ConceptNet \cite{Havasietal:2009,Speeretal:2012} is a semantic network that contains commonsense facts about the world collected from volunteers over the Internet.  %(via templates, free text, games etc.). 
Nodes in the network represent concepts in the form of words or phrases, % in natural language, 
and edges the knowledge relations holding between them (e.g.,\ \textit{health insurance }\textsc{CapableOf} \textit{cover ambulance transportation}). 
The inventory covers 37 relations, some of which are commonly used in other resources like WordNet (e.g.,\ \textsc{IsA, PartOf}) while most others are targeted for capturing commonsense knowledge and as such are particular to ConceptNet (e.g.,\ \textsc{HasPrerequisite, MotivatedByGoal}). 
\\The annotation was performed by two annotators in parallel who were asked to label all argumentative units and inserted sentences with ConceptNet relations (irrespective of whether or not the relation {\em instance} is covered in ConceptNet). %This annotation was performed for each
%, whereby each 
%argumentative unit/inserted sentence
%was labelled 
%separately. 
The annotators labeled the complete relation triple by (I) selecting and marking two concepts (from the same argumentative unit/inserted sentence), and (II) the ConceptNet relation that they judge to hold between them. %, using the relation inventory of ConceptNet. 
Two examples from our dataset are given in  Fig.\ \ref{fig:relation-examples}. Note that we didn't mark the concepts beforehand, but let our annotators label both: the concepts %within the sentence 
\textit{and} the relation between them.\footnote{This sometimes led to disagreements between annotators regarding the span they selected for the same concept, e.g. for the sentence \textit{Sophisticated programmes should be financed by the licence fee}, A annotated the triple  \textit{sophisticated programme, financed by licence fee} \textsc{(ReceivesAction)}, and B  \textit{sophisticated programme, financed} \textsc{(ReceivesAction)}. This disagreement was harmonized in the gold version by the expert annotator.}

%\orangenote{[Q: Will you ever collect annotations between units, especially overt argument units and implicit statements? Aren't they important? If you don't need it, you should say how the annotations connect the explicit and implicit text (by analyzing the concept overlap of both regions)--> I put it in the outlook}

\begin{figure}[t]
\begin{centering}
\includegraphics[width=0.48
\textwidth]{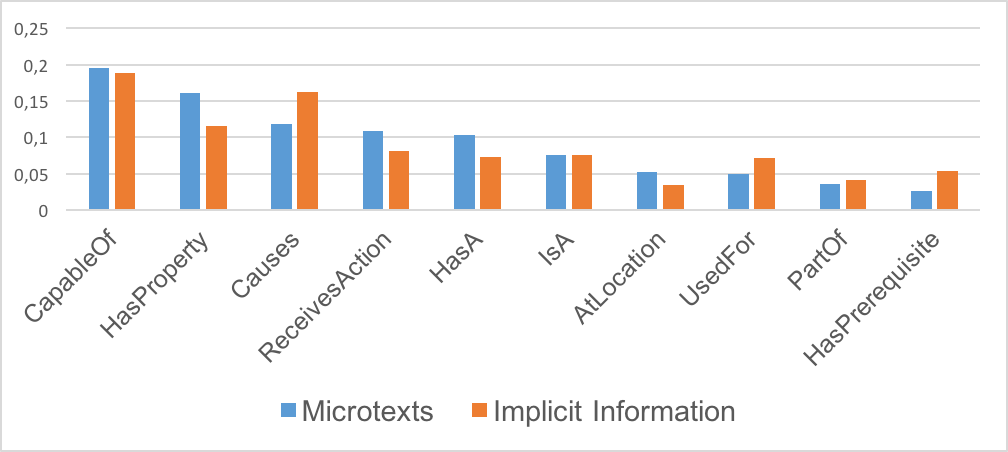}
\caption{Distribution of ConceptNet relations within Microtexts and Implicit Information Annotations (in \%)}
\label{fig:CN_dis}
\vspace{0.6cm}
\end{centering}
\end{figure}

%\greennote{maybe also give an example where the relation bridges two sentences; for this further example: choose some more ConceptNet-specific relations}\\ 
In preliminary annotation experiments we observed that in many cases several relations can be suitable for the same sentence (cf.\ example 2 in Fig.\ \ref{fig:relation-examples}), therefore we allowed for more than one relation per sentence/argumentative unit.
%\greennote{Now I'm confused: i thought you talk about several options for a given pair of concepts. But you really mean "per sentence". This raises a question about how the task was presented to the annotators: did they receive sentences with pre-labeled concepts and needed to label each possible pairing with relation types?}
If none of the relations covered by the ConceptNet relation inventory was fitting, our annotators inserted \textit{NONE} and collected suggestions for additional relations (such as \textsc{Requires}). %, which is not captured by the existing ConceptNet inventory). 
We release these suggestions along with examples from our data together with our dataset. \\

\begin{figure}[t]
\small{(I) \textit{Fees result in longer durations of studies.}} \\
Annotation: \textit{fees, longer durations of studies} (\textsc{Causes})\\\\
(II) \textit{Dog dirt is disgusting and a hygiene problem}. \\
Annotation: \textit{dog dirt, disgusting} (\textsc{HasProperty})/
\\\noindent\hspace*{16mm}\textit{dog dirt, hygiene problem}
(\textsc{IsA})
\vspace{-0.25cm}
\caption{Sentences from our dataset annotated with ConceptNet relations.}\label{fig:relation-examples}

\label{fig:cn-examples}
\end{figure}

\textbf{Statistics.}
Our annotators used 25 relations from the inventory of 37 relation types provided by ConceptNet to label the argumentative units from the Microtexts and the inserted sentences. We measure annotator agreement for (I) the marked concepts in order to evaluate if our annotators agree on the spans of texts selected as concepts, and for (II) the assigned relations (separately). (I) we measure in terms of word overlap and obtain high averaged word overlap scores of 76.98\% (Jaccard) and 84.87\% (Dice), indicating solid agreement between the selected concepts. (II) we measure in terms of Cohen's Kappa and achieve a moderate agreement of 45.05\%. We produce a gold standard done by an expert annotator (one of the authors) which provides the basis of our final analysis. In this gold standard, on average 3.58 relation triples were assigned per argumentative unit and 3.01 relation triples per inserted sentence. The distribution of the 10 most frequent relation types is shown in Fig.\ \ref{fig:CN_dis}.

The most frequently occurring relation is \textsc{CapableOf} (19\% within argumentative units and 20\% within inserted sentences) followed by \textsc{HasProperty} (16/12\%) and \textsc{Causes} (12/16\%). %\textsc{ReceivesAction} (11/8\%),  and \textsc{HasA} (10/7\%). 
The largest differences between relations assigned to argumentative unit vs.\ inserted sentence we observe for \textsc{HasProperty} (4.6pp) and \textsc{ReceivesAction} (2.8pp), both more prominent in microtexts, and \textsc{Causes} (4.4pp), more prominent in implicit knowledge annotations. %\orangenote{[which ones more prominent in which text type?]}
We find that only 9 of 576 argumentative units (1.56\%) and 24 of 1295 inserted sentences (1.85\%) were identified as cases where none of the relations covered by the relation inventory fits, which %. The high amount of sentences which can be mapped to commonsense knowledge relations 
points to the fact that knowledge repositories such as ConceptNet can play an important role in 
 argument analysis and the retrieval of implicit knowledge.

%, while there is a relatively high amount of rarely annotated relation types.

\section{Analysis of the Annotations: Visualizing Correlations}

In this section, we analyse correlations between the labels and properties annotated for our dataset. In addition to the analysis of the statistics and distribution of the labels annotated in our corpus (cf. Sec.\ 4), we want to reveal patterns and intersections between the annotation categories and properties of our dataset, with the goal of learning more about the characteristics of both argumentative texts and implicit knowledge in terms of structural features and semantic information. We expect the results of our analysis to be helpful for guiding and enhancing the process of automated argument analysis as well as of the automatic reconstruction of implicit knowledge in argumentative texts. 

\subsection{Number of Hops} 

\textbf{Hops - Adjacency and Relatedness of Argument Units.} The gold version of our dataset contains 719 pairs of argumentative units. %Out of these, 464 pairs are adjacent while 255 are not. Further,  462 pairs stand in a direct argumentative relation while 257 do not.
1295 sentences were inserted, that is on average 1.8 sentences per argument pair. The pairs of argumentative units either stand in a direct argumentative relation, which means that they are directly connected in the argument graph (like %the pairs 
e$_1$ and e$_2$ %, or e$_1$ and e$_5$ 
in Fig.\ \ref{fig:arggraph}), or the units are adjacent to each other (e$_1$ and e$_2$, e$_2$ and e$_3$ ...), or both (e$_1$ and e$_2$). We expect that more inserted sentences are needed to connect argument pairs that stand in an argumentative relation but 
are are not adjacent, since the missing information could be included in the intermediate argument units (e.g., what is missing between e$_1$ and e$_5$ in Fig.\ \ref{fig:arggraph} could be expressed in e$_2$, e$_3$ or e$_4$). We also hypothesize that more implicit information is needed to connect argument pairs that don't stand in a direct argumentative relation, %but are adjacent%(like e2 and e3 in Fig. \ref{fig:arggraph})
since argument units that aren't related can come from different chains of the argument and might therefore require more explications % of implicit knowledge  
than directly related argument units
(cf.\ Fig.\ \ref{fig:arggraph}, e$_4$ and e$_5$). %in Fig.\ \ref{fig:arggraph}).  %\orangenote{[Not sure I share this intuition. If two sentences are next to each other, they should be coherent, at least. If they are adjacent, but not in an argumentative relation, I don't think you will have much to say about what might connect them (since there is no argumentative link)? -- but you better see what happens. But claiming that adjacent sentences are not coherent I think is dangerous.]}
%\orangenote{[AF: little to nothing has been said by now about what it really means that two AUs in a relation are not adjacent. It would be worthwhile to show an example (of both), so that the reader can better judge what differences to expect for the different cases. If there is no hypothesis, why should we care? - could go into the motivation]}
Since -- by our annotation design -- the inserted sentences contain the minimal amount of information that makes the connection between two argumentative units explicit, we interpret each inserted sentence as one hop that is needed to connect the given argument pair.%\footnote{Recall that annotators were asked to express the missing content in simple statements, ideally covering single atomic propositions.}  
\\We find only a relatively small difference in the average number of sentences inserted between adjacent (1.9) and non-adjacent units (1.62) (cf.\ Table \ref{tab:generalstats}), indicating that it is not the case that more hops (inserted sentences) are needed when units are not adjacent.  Interestingly, on the other hand we observe a remarkable difference between the number of sentences inserted between argumentatively related (1.6, Table \ref{tab:corr_argrel_hops}) and non-related units (2.14, Table \ref{tab:corr_argrel_hops}). This indicates that more hops are needed when there is no direct argumentative relation between the argument units.

\begin{table}
\begin{center}
\begin{tabular}{ c c c } 
 \toprule
 & adjacent & not adjacent \\ 
 \midrule
 nb. of pairs & 464 & 255 \\  percentage & 0.65 & 0.35 \\ 
 nb. of inserted sentences & 881 & 414 \\ 
 inserted sentences (avg) & 1.9 & 1.62 \\ 
\bottomrule
\end{tabular}
\caption{Adjacency of argument pairs and number of inserted sentences, Gold. %Inserted sentences are given as average number of sentences per pair.
}
\label{tab:generalstats}
\end{center}
\end{table}

% \orangenote{[Please try to state this in a more compressed way; there is also an ambiguity: do you talk about the number of relations that receive (any number of) inserted sentences; or about how many inserted sentences in average we count across the different types of argumentative relations?]}
%, e.g. most sentences are inserted by our annotators between argumentative units that stand in an \textit{undercut} relation (1.84 sentences inserted on average), while between argumentative units which relation is labelled as example only 1.11 sentences are inserted on average. 
%While we could only find a minor difference between the number of inserted sentences inserted between adjacent and non-adjacent units, 

\textbf{Hops - Argumentative Relations.} After revealing that more hops are required for connecting non-related argumentative units, % that do not stand in a direct argumentative relation, 
we are also interested whether there are argumentative relations for which more hops are needed then for others.
Our dataset contains 5 argumentative relations, with \textit{support} being the most frequent one (37\%) followed by \textit{rebuttal} (15\%) and \textit{undercut} (8\%) (cf. Table \ref{tab:corr_argrel_hops}). %\orangenote{[To help readers not familiar with argumentation, you should mention who designed this scheme, you can also illustrate their meaning in the example of argumentation graphs mentioned above; please also provide the definitions in the appendix.]MB: this is described in Sec 3.2, and I added an example}
We find that for the \textit{undercut} relation, most sentences are inserted on average (1.84). %sentences inserted on average). 
This makes sense since \textit{undercuts} challenge the acceptability of an inference between two propositions (cf.\ \newcite{peld:stede:2015}) and can therefore be seen as a very complex %argumentative 
relation that requires more explications than others. The least sentences are inserted on average for \textit{example} relations (1.11 avg.), indicating that these are relations that usually don't need multi-hop connections of implicit knowledge.

\begin{table}
\begin{tabular}{@{}c@{~~}cc|cc@{}}
\toprule
& \multicolumn{2}{l}{argument relation} & \multicolumn{2}{l}{inserted sentences} \\
 %     \hline
        & total &percentage &  total & per relation \\
\midrule
%\hline       \hline
support  & 263 &37& 423& 1.61\\
rebuttal & 108 &15& 165& 1.53\\
undercut & 61 &8  & 112& 1.84\\
addition & 21 &3  & 34 & 1.62\\
example  & 9 &1   & 10 & 1.11\\
      \midrule
      relations total & 462 &100 &  744 & 1.6\\
      \midrule
      non-related units & 257 &36 &  551 & 2.14\\    \midrule \midrule %\hline \hline
TOTAL    & 719 &100 & 1295& 1.8\\
\bottomrule
\end{tabular}
    \caption{Correlation between argumentative relations and number of hops (inserted sentences).}
\label{tab:corr_argrel_hops}
\vspace{0.3cm}
\end{table}

\begin{figure}[t]
\begin{centering}
\includegraphics[width=0.47
\textwidth]{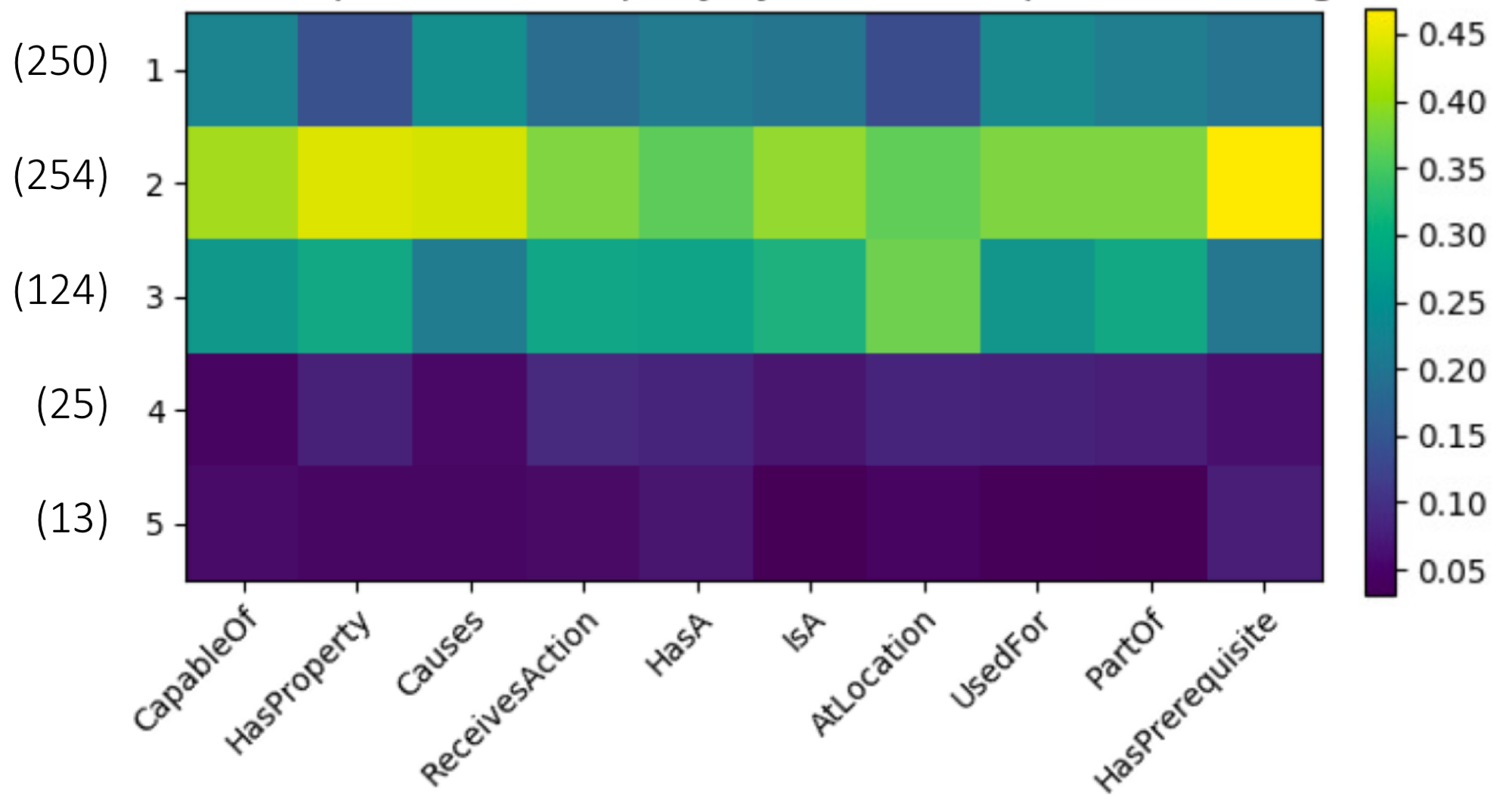}
\vspace{-0.3cm}
\caption{Distribution of commonsense relations within inserted sentences among hops, rel. freq. by relation type, with total number of hops given on the left side.} %Left: relative frequency by relation type, right: relative frequency by number of hops.}
\vspace{0.3cm}
\label{fig:hops-cn}
\end{centering}
\end{figure}

\textbf{Hops - Commonsense Relations.}
Next, we are interested whether there are co-occurrences between the number of hops and commonsense relation types. We want to investigate whether specific commonsense relation types appear more often in single (one inserted sentence) vs. multiple hops (more than one inserted sentence). Therefore, for all commonsense relations within inserted sentences, we count how often they occur in one hop connections (when one sentence was inserted as missing information), in two hop connections and so on. The resulting heatmap %(10 most frequent relations) 
is displayed in Fig. \ref{fig:hops-cn}. %\textcolor{red}{(left) there is no left Figure in 5!}. 
We observe that all relations occur most often within a set of two inserted sentences, % (two hops), 
which  corresponds to the average number of inserted sentences (1.8, cf. Table \ref{tab:corr_argrel_hops}). Interestingly, \textsc{HasProperty} and \textsc{AtLocation} are relations which occur only rarely within one hop connections, the latter being most often used in sets of three inserted sentences. Those relations seem to mark information units that require other pieces of information to connect an argument pair. 
%Fig. \ref{fig:hops-cn} (left) shows the distribution of commonsense relations for each number of hops 
%the distribution of commonsense relations within inserted sentences.

%\orangenote{What about frequent uni- (covered in Fig.2), bi- or tri-grams?}

%\orangenote{We do not have a distribution of the number/length of inserted sentences IRRESPECTIVE of the CSRs. If it is mostly 2, then what we see is not surprising (except for a few outliers. Maybe show a heatmap over the overall 1, 2, 3, 4, 5 statement sets alongside?}

\begin{figure}[t]
\begin{centering}
\includegraphics[width=0.23
\textwidth]{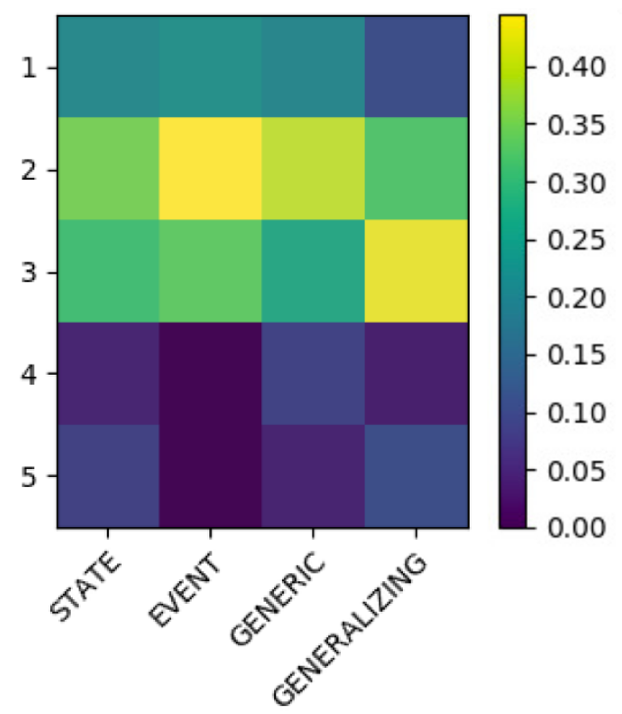}
\vspace{-0.4cm}
\caption{Distribution of Semantic Clause Types among hops, relative frequency by clause type.}%, right: relative frequency by number of hops.}
\vspace{0.9cm}
\label{fig:hops-se}
\end{centering}
\end{figure}

\textbf{Hops - Semantic Clause Types.} Similarly, we want to investigate co-occurrences between the number of hops and semantic clause types. Again, for all clause types within inserted sentences, we count how often they occur in each set of inserted sentences (1-5),  Fig. \ref{fig:hops-se} shows the resulting heatmap. We find that \textsc{States, Events} and \textsc{Generic Sentences} occur most often within two hop connections, while \textsc{Generalizing Sentences} are most often used within sets of three inserted sentences and rarely when only one sentence was inserted. \textsc{Generalizing Sentences} therefore can be interpreted as markers of information units that stand-alone are not able to connect argument pairs, but rather co-occur with other pieces of information for filling knowledge gaps in argumentative texts.

\subsection{Adjacency and Argumentative Relatedness}

When filling knowledge gaps in argumentative texts automatically, it might be useful to leverage the structure of an argument and to determine which type of knowledge exactly is missing for which pair of argument units. Knowing the semantic properties of the knowledge that is needed to connect argument units that are -- for example -- adjacent vs.\ those that are not, can guide the process of extracting knowledge for filling these gaps. Therefore, we want to investigate whether the distribution of the semantic properties we annotated for the inserted sentences -- commonsense relation types and semantic clause types --  respectively differs depending on the internal structure of an argument, in our case this is whether (i) the arguments from a given pair are adjacent or not, and/or whether (ii) the arguments from a given pair are argumentatively related or not. 

\begin{figure}[t]
\begin{centering}
\includegraphics[width=0.48
\textwidth]{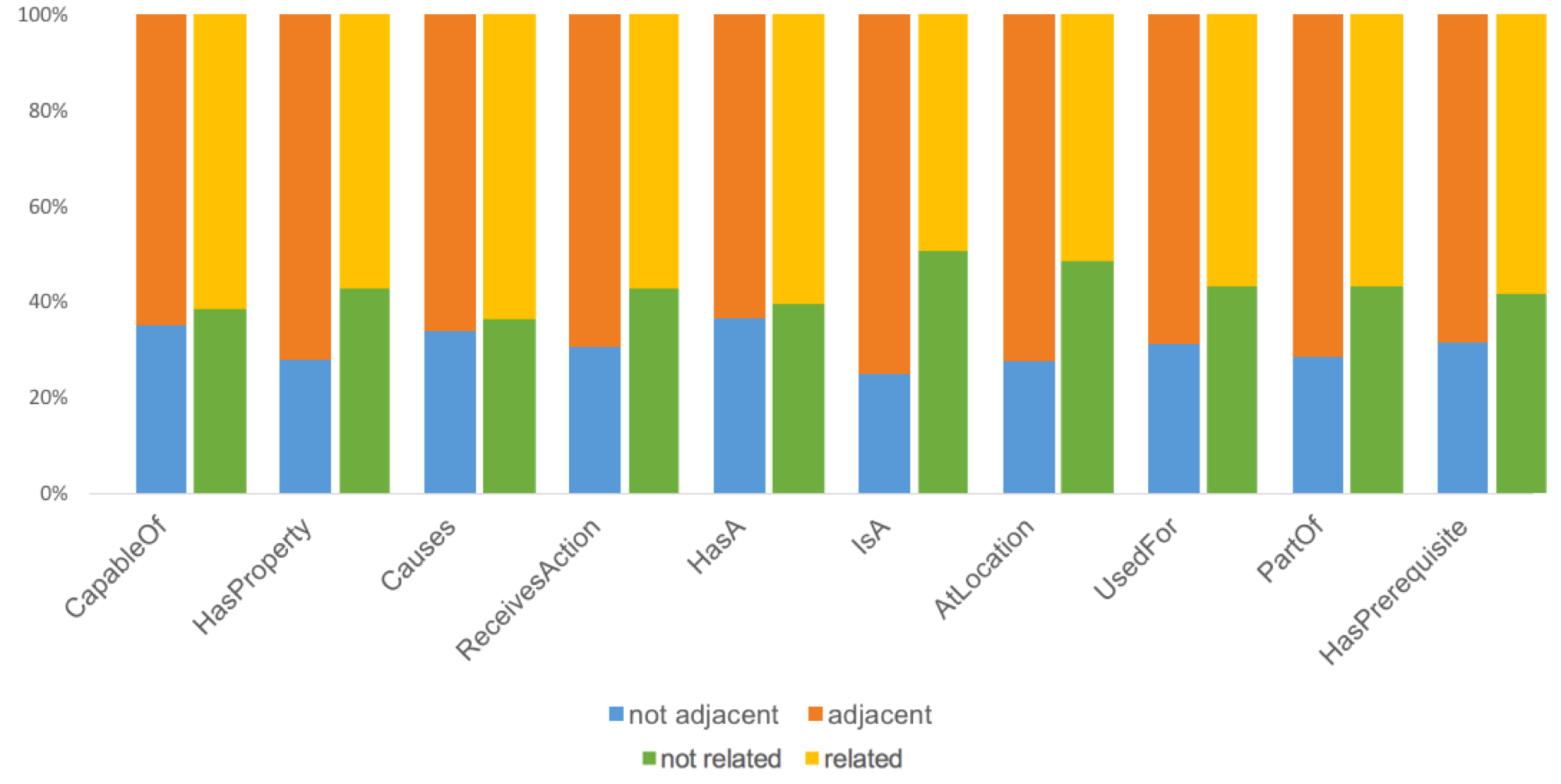}
\vspace{-0.7cm}
\caption{Adjacency and Argumentative Relatedness for Commonsense Relations (in \%).}
\vspace{0.3cm}
\label{fig:adjacency-rel-cn}
\end{centering}
\end{figure}

\textbf{Commonsense Relations.} Fig.\ \ref{fig:adjacency-rel-cn} (blue/orange bars) shows that the distribution of commonsense relation types only slightly differs between adjacent and non-adjacent units. We find that % the relations
\textsc{IsA} (75\%), \textsc{AtLocation} and \textsc{HasProperty} (both 72\%) occur most often within sentences inserted between adjacent units, while \textsc{HasA} and \textsc{CapableOf} are relations that occur more often in sentences inserted between non-adjacent units (36\% and 35\%). % for non-adjacent units).
We also observe only slight variations regarding the distribution of commonsense relations between units that are argumentatively related and those that are not (Fig.\ \ref{fig:adjacency-rel-cn}, green/yellow bars). While \textsc{Causes} (64\%), \textsc{CapableOf} (61\%) and \textsc{HasA} (61\%) are often assigned to sentences inserted between related units and therefore can be interpreted as argumentatively relevant, \textsc{IsA}	and \textsc{AtLocation} are typical labels for implicit information between units that don't stand in an direct argumentative relation (51\% and 49\% for not related units). 
%\orangenote{This is nice: Causes and CapableOf are clearly more argumentatively relevant. I suggest that you mention those that cover more than some suitable cut: 50 or 60\%: there is also Has Prer. }

\begin{figure}[t]
\begin{centering}
\includegraphics[width=0.4
\textwidth]{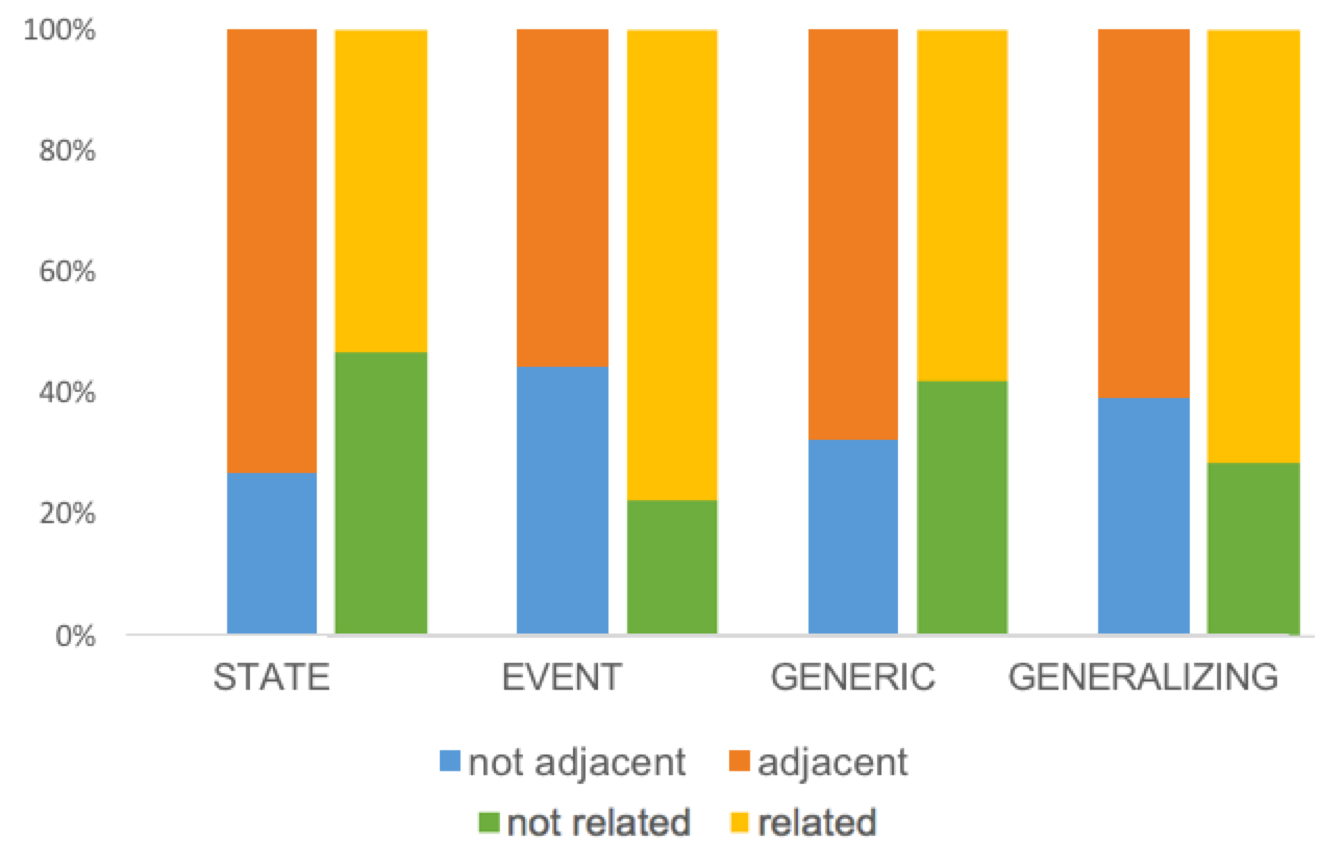}
\vspace{-0.4cm}
\caption{Adjacency and Argumentative Relatedness for Semantic Clause Types (in \%).}
\vspace{0.3cm}
\label{fig:adjacency-rel-se}
\end{centering}
\end{figure}

\textbf{Semantic Clause Types.} We also want to investigate whether the distribution of  Semantic Clause Types differs between adjacent and not adjacent units, and/or between argumentatively related and not related units. Fig.\ \ref{fig:adjacency-rel-se} (blue/orange bars) shows that \textsc{States} occur most often between units that are adjacent (73\%), while \textsc{Events} show the lowest proportion for adjacent units (56\%). Regarding the distribution of semantic clause types assigned to sentences between argumentatively related and not related units (Fig.\ \ref{fig:adjacency-rel-se}, green/yellow bars), we find a large difference for \textsc{Events} (78\% between related and 22\% between not related units) and \textsc{Generalizing Sentences} (71\%/29\%), while \textsc{States} (53\%/47\%) and \textsc{Generic Sentences} (58\%/42\%) are more equally distributed.%\footnote{For the correlation between adjacency and argumentative relatedness, both labels taken from the Microtext Corpus, cf. the attachment distance analysis in \newcite{peld:stede:2015}.} 

\subsection{Correlations between Assigned Labels}

In this section, we analyse correlations between argumentative relations, commonsense relations and semantic clause types. We want to reveal patterns and intersections between the annotation categories in order to learn more about the structural features and semantic properties of both argumentative texts and implicit knowledge. For all analyses reported in this section, we measure correlations using the Matthews correlation coefficient (MCC) \cite{mcc}, which assigns correlation coefficient values between -1 and +1 to pairs of labels (here e.g. \textit{support} and \textsc{Causes}). A coefficient of +1 represents a perfect correlation, 0 an average random prediction, and -1 an inverse correlation. 

\begin{figure}[t]
\begin{centering}
\includegraphics[width=0.48
\textwidth]{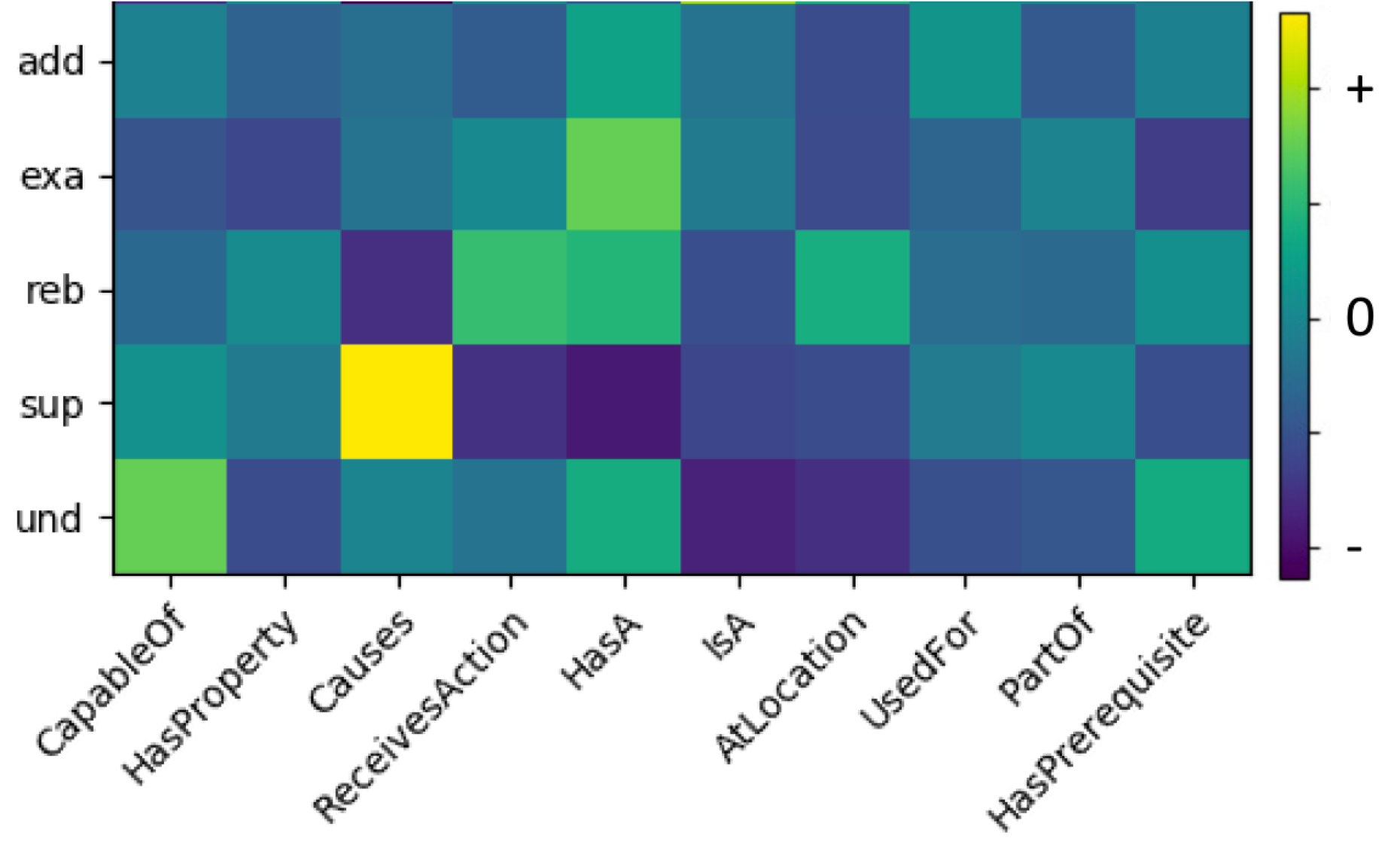}
\vspace{-0.4cm}
\caption{Correlations between Argumentative Relations and Commonsense Relations, MCC correlation matrix. Bright colours indicate positive and dark colours negative correlation from a scale between +1 and -1.}
%, right: relative frequency by commonsense relation.}
\vspace{0.3cm}
\label{fig:argrel-cn}
\end{centering}
\end{figure}

\textbf{Argumentative Relation - Commonsense Relation.} 
First, we look at correlations between argumentative relations and commonsense relations. We want to investigate if specific commonsense relations express specific argumentative relations, and if specific argumentative relations are more characteristic for specific commonsense relations than others. Fig.\ \ref{fig:argrel-cn} shows that the relation \textsc{Causes} is very dominant within sentences inserted between argument units that stand in a \textit{support} relation, which reveals the importance of causal explanations for filling knowledge gaps between supporting argument units. An example of our dataset is given in Fig.\ \ref{fig:sup-causes-example}. The relations \textsc{ReceivesAction} and \textsc{HasA} correlate negatively with \textit{support} but positively with \textit{rebuttals}, underlining the difference in distributions of commonsense relation types between these two contrary argumentative relations. We also observe that \textit{rebuttals} correlate negatively with \textsc{Causes}, indicating that causal explanations are not typical for connecting argument units that rebut each other.

\begin{figure}[t]
\small{(e$_2$) \textit{The developments in that conflict should not be left} 
\\\noindent\hspace*{5.5mm} \textit{to former Cold War opponents alone,}}\\
(e$_3$) \textit{for that course can only lead to escalation in some form.}
\\
--------------------------------------------------------------------\\
Implicit Information: \textit{A conflict may lead to escalation}. \\
Commonsense Relation: \textit{conflict, escalation} (\textsc{Causes})\\\\
\vspace{-0.9cm}
\caption{Example of a causal explication for a support relation (e$_3$ supports e$_2$).}
\label{fig:sup-causes-example}
\end{figure}

%\orangenote{Very nice! This deserves an example that makes this very transparent to the reader!}
%\orangenote{There are more that seem to make sense: CapableOf/HasPrerequisite - Undercut; HasA - Example; IsA - inverse for undercut; HasPrerequisite for Example; maybe also: HasA for addition }

%\bluenote{Are specific argumentative relations more characteristic for specific commonsense relations? Do specific commonsense relations express specific argumentative relations? Are specific commonsense relations typical for certain argumentative relations?}

\textbf{Argumentative Relation - Semantic Clause Type.} Next, we are interested in the correlations between argumentative relations and semantic clause types. We analyse if specific argumentative relations are more characteristic for specific clause types, and vice versa, if specific clause types correlate with specific argumentative relations. Fig.\ \ref{fig:argrel-cn} shows that \textit{examples} differ from the rest of argumentative relations regarding the correlations with clause types: while we find high positive correlations with \textsc{States} and \textsc{Events}, \textsc{Generics} very infrequently co-occur with \textit{examples}. This makes sense since examples usually express knowledge about individuals rather then generic knowledge (cf. \newcite{beckeretal:2016a}). %, \orangenote{and the same seems to apply for inserted sentences that explain a connection between argument units that stand in an \textit{example} relation [you are repeating yourself]}. 
Our correlation analysis also reveals interesting patterns regarding the \textit{support} relation: here we find a negative correlation with \textsc{States} and a positive correlation with \textsc{Generic Sentences}, indicating the importance of generic knowledge for sentences that connect two argument units which support each other. Interestingly, when looking at the correlations between \textsc{Generic Sentences} and argument relations, we find that this is the only positive correlation, while all others are negative. This underlines the finding that \textsc{Generics} can be seen as an important feature of sentences inserted between supporting argument units. 

%\orangenote{addition shows some correlation and also undercut, but weaker. Which color is 0? %I will add legends to the plots}

\begin{figure}[t]
\begin{centering}
\includegraphics[width=0.25
\textwidth]{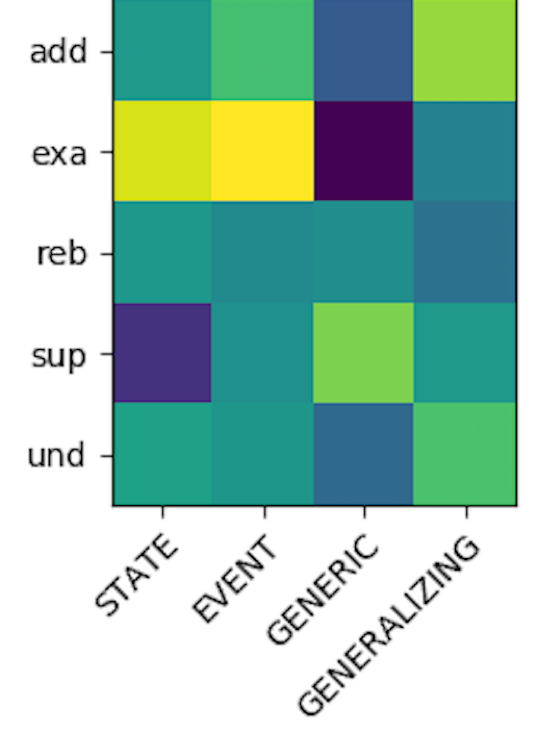}
\vspace{-0.4cm}
\caption{Correlations between Argumentative Relations and Semantic Clause Types, MCC correlation matrix.}
\vspace{0.3cm}
\label{fig:argrel-se}
\end{centering}
\end{figure}

\begin{figure}[t]
\begin{centering}
\includegraphics[width=0.4
\textwidth]{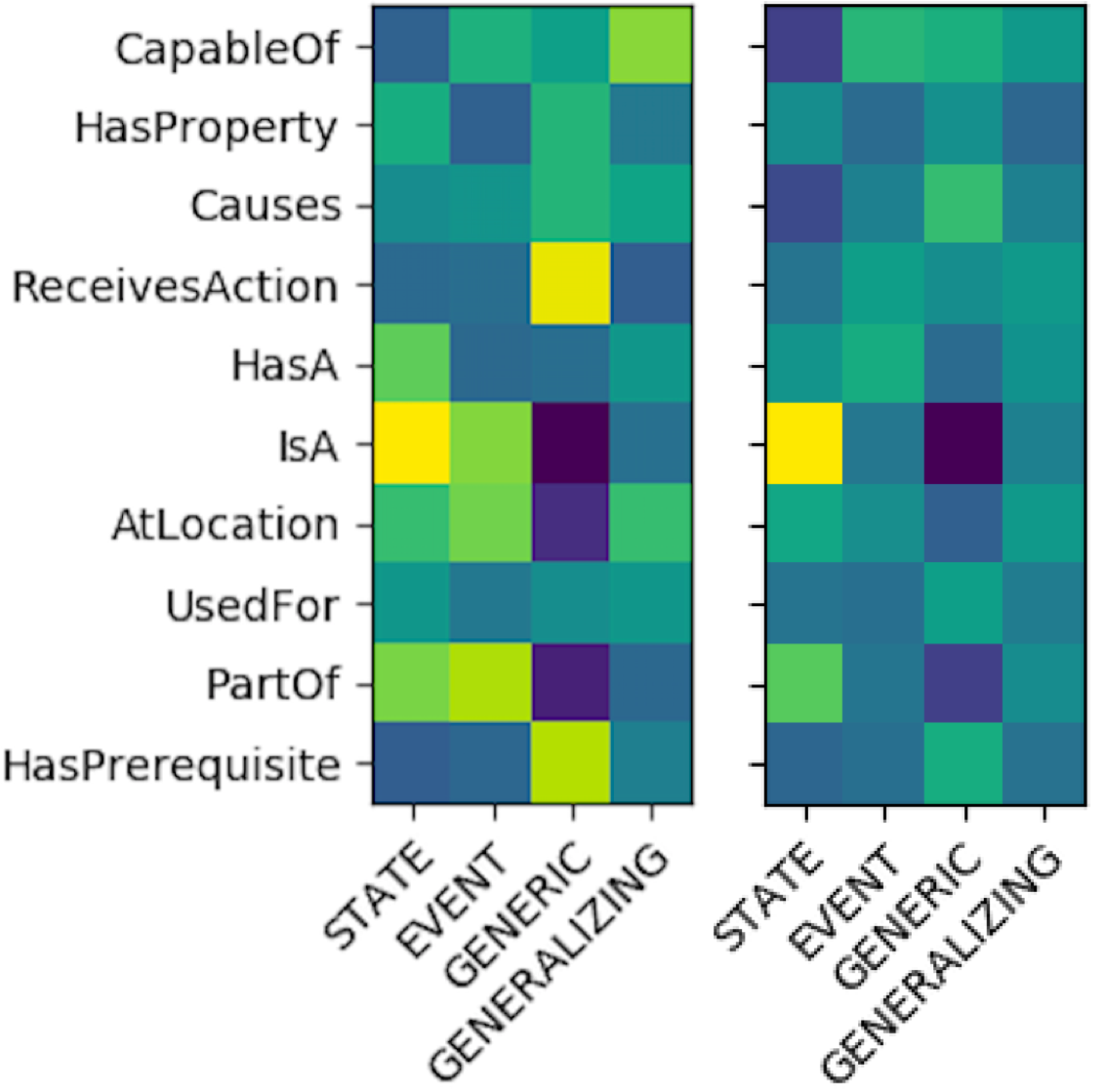}
\vspace{-0.3cm}
\caption{Correlations between Commonsense Relations and Semantic Clause Types in Microtexts (left) vs. Inserted Sentences (right), MCC correlation matrix.}
\vspace{0.3cm}
\label{fig:cn-se}
\end{centering}
\end{figure}

\textbf{Commonsense Relation - Semantic Clause Type.} 
There are also some interesting correlations between commonsense relations and semantic clause types which we display in Fig.\ \ref{fig:cn-se}. We aim to discover whether (i) specific clause types are indicators for certain commonsense relations or vice versa, and whether (ii) the distribution of clause types among certain commonsense relations differs between microtexts and inserted sentences. 
Fig.\ \ref{fig:cn-se} (left) shows that within the microtexts, \textsc{Generic Sentences} correlate negatively with \textsc{IsA, AtLocation} and \textsc{PartOf}, and positively with \textsc{ReceivesAction} and \textsc{HasPrerequisite}. For the three relations \textsc{IsA, AtLocation} and \textsc{PartOf} we find a positive correlation with \textsc{States} and \textsc{Events} and a negative correlation with \textsc{Generic Sentences}, indicating that these relations typically express individual rather than generic knowledge.
Fig.\ \ref{fig:cn-se} (right) shows that the correlations within the inserted sentences are not as strong as in the microtexts, but still we can see that similar to the microtexts, \textsc{Generic Sentences} correlate negatively with \textsc{IsA} and \textsc{PartOf}, while we can't observe a strong positive correlation between \textsc{States} and any commonsense relation. The only high positive correlation we find is between \textsc{IsA} and \textsc{States}, indicating that within the inserted sentences (as well as within the microtexts), \textsc{IsA} relations typically describe specific properties of individuals (cf. Sec.\ 4.1).

\section{Conclusion and Outlook}
In this paper, we presented a carefully designed dataset consisting of high-quality human annotations of implicit knowledge in argumentative texts. To learn more about the characteristics of both the argumentative texts and the added information, we further annotated the data with semantic clause types and commonsense knowledge relations. 
We then provided an in-depth analysis of our annotated dataset with the goal of  revealing characteristic distributions and correlations, co-occurring patterns and intersections between the annotation categories. % and properties. 
This helped us to gain insights into the properties of both argumentative texts and implicit knowledge in terms of structural features and semantic information: We found for example that \textsc{Generic Sentences} play a dominant role within the inserted sentences, % added by our annotators, 
indicating the relevance of generic knowledge within implicit information. Almost all sentences in our dataset -- from both the microtexts and the inserted information -- could be mapped to commonsense knowledge relations, pointing to the fact that knowledge repositories such as ConceptNet can play an important role in argument analysis and are an important source for the retrieval of implicit knowledge. \\
When analyzing correlations between the labels and structural properties of our dataset, we could furthermore reveal patterns and intersections between the annotation categories and structures of our dataset: % and, built on that, gain insights into the properties of both argumentative texts and implicit knowledge.% in terms of structural features and semantic information. 
We found for example that more inserted sentences are needed when there is no direct argumentative relation between the argumentative units, and that complex argumentative relations such as \textit{undercut} require more explications than other relations.
Our correlation analysis further demonstrated the benefit of leveraging the structure of an argument and the characterization of the type of knowledge that is needed to connect argument pairs. We investigated whether the distribution of the semantic properties we annotated for the inserted sentences differs depending on the internal structure of an argument and revealed for example that \textsc{States} occur most often between units that are adjacent, while \textsc{Events} are frequently used for connecting argumentatively related units. \\
Finally, when investigating correlations between argumentative relations, commonsense relations and semantic clause types, we could for example reveal the importance of causal explanations (as expressed by the relation \textsc{Causes}) for filling knowledge gaps between supporting argument units. \textsc{Generics} also turned out to be a important feature of sentences inserted between supporting argument units.

%%Finally, we wanted to learn more about the properties of the texts in our dataset in terms of co-occurrences of commonsense relation types, and interpreted the observed co-occurrences such as \textsc{HasPrerequisite} and \textsc{HasProperty}, or \textsc{CapableOf} and \textsc{Causes} as features of the text types argumentative texts and of implicit knowledge annotations, respectively.

The knowledge we gained about the properties of argumentative texts and implicit knowledge, and our observations on their interaction can assist automated argument analysis, e.g.,\ it can be beneficial for assessing the strength of an argument, apart from the benefit of making the underlying logics of the argument transparent for both humans and computational systems. The results from our in-depth analysis %of the characteristics of both argumentative texts and implicit information and their intersections and connections 
can furthermore guide the process of revealing implicit information in argumentative texts automatically, e.g.\ by utilizing the revealed properties of implicit information and the observed relations between implicit information and the surrounding argument units.

%\orangenote{This sounds a bit as if you were at the beginning of this examination, not at the end.. Once you have worked out some solid insights as conclusions, you can use some of this as motivation for the investigation.}

We release our dataset as an extension to the Microtext corpus. We expect it to be a useful starting point for automatically filling knowledge gaps in arguments, and we hope that it will inspire future research on argument analysis and implicit knowledge acquisition.

\begin{comment}
\begin{itemize}
\itemsep-5pt
\item refer to my own work
\item list use cases for this dataset/ideas for utilizing the annotations: verbalization of CN relations in texts (find indicators manually, calculate how often they occur, examples…), suggest for which downstream task SE types/CN relations annotations can be useful
\item automatic analyses
\item challenges
\item additional interesting annotation categories: frames etc.
\item summarize results and contributions, outlook
\item  collect cn annotations between units, especially overt argument units and implicit statements
\end{itemize}
\end{comment}

%\section{Acknowledgements}
%This work has been funded by the Deutsche Forschungsgemeinschaft (DFG)
%within the project ExpLAIN, Grant Number STU 266/14-1 and FR 1707/-4-1,
%as part of the Priority Program “Robust Argumentation Machines (RATIO)” (SPP-1999). We thank our annotators Anna Becker and Katharina Korfhage for their contribution.

%\section{References}
\bibliographystyle{acl_natbib}
\bibliography{acl2020}
%\bibliographystylelanguageresource{lrec}
%\bibliographylanguageresource{languageresource}
%\section*{Appendix}

\end{document}